\documentclass[conference]{IEEEtran}
\usepackage[absolute,showboxes,overlay]{textpos}

\TPMargin{5pt}

\newcommand{\copyrightstatement}{
    \begin{textblock}{0.84}(0.08,0.93)    
         \noindent
         \footnotesize
         \copyright 2021 IEEE. Personal use of this material is permitted. Permission from IEEE must be obtained for all other uses, in any current or future media, including reprinting/republishing this material for advertising or promotional purposes, creating new collective works, for resale or redistribution to servers or lists, or reuse of any copyrighted component of this work in other works. {Cite from IEEE: }{DOI No. 10.1109/GCWkshps52748.2021.9681832}
    \end{textblock}
} 

\IEEEoverridecommandlockouts
\usepackage{cite}
\usepackage{amsmath,amssymb,amsfonts}
\usepackage{graphicx}
\usepackage{textcomp}
\usepackage{xcolor}

\usepackage{subfigure}
\usepackage{fancyhdr}
\usepackage{algorithm}
\usepackage{algpseudocode}
\usepackage{booktabs} 
\usepackage{multirow}
\usepackage[acronym]{glossaries}

\newacronym{rf}{RF}{Radio frequency}
\newacronym{sri}{SRI}{Soft range information}
\newacronym{nlos}{NLOS}{non-line-of-sight}
\newacronym{los}{LOS}{line-of-sight}
\newacronym{5g}{5G}{fifth generation}
\newacronym{de}{DE}{distance estimate}
\newacronym{ml}{ML}{Machine learning}
\newacronym{dl}{DL}{Deep learning}
\newacronym{uwb}{UWB}{Ultra-wideband}

\include{pythonlisting}

\def\BibTeX{{\rm B\kern-.05em{\sc i\kern-.025em b}\kern-.08em
    T\kern-.1667em\lower.7ex\hbox{E}\kern-.125emX}}
\begin{document}
\copyrightstatement
\title{A Deep Learning Approach for Generating Soft Range Information from RF Data}

\author{
\IEEEauthorblockN{
Yuxiao~Li\IEEEauthorrefmark{1},
Santiago~Mazuelas\IEEEauthorrefmark{2}, and
Yuan~Shen\IEEEauthorrefmark{1}}
\IEEEauthorblockA{\IEEEauthorrefmark{1}
Department of Electronic Engineering,
Tsinghua University,
Beijing, China \\
}
\IEEEauthorblockA{\IEEEauthorrefmark{2}
BCAM-Basque Center for Applied Mathematics, and IKERBASQUE-Basque Foundation for Science, Bilbao, Spain \\
Email: li-yx18@mails.tsinghua.edu.cn,
smazuelas@bcamath.org,
shenyuan\_ee@tsinghua.edu.cn
}}

\maketitle

\begin{abstract}

\gls{rf}-based techniques are widely adopted for indoor localization despite the challenges in extracting sufficient information from measurements. \Gls{sri} offers a promising alternative for highly accurate localization that  gives all probable range values rather than a single estimate of distance. We propose a deep learning approach to generate accurate \gls{sri} from \gls{rf} measurements. In particular, the proposed approach is implemented by a network with two neural modules and conducts the generation directly from raw data. Extensive experiments on a case study with two public datasets are conducted to quantify the efficiency in different indoor localization tasks. The results show that the proposed approach can generate highly accurate \gls{sri}, and significantly outperforms conventional techniques in both \gls{nlos} detection and ranging error mitigation.

\end{abstract}


\section{Introduction}
\label{sec:intro}

Positional information is a key enabler for the \gls{5g} of mobile communications and beyond, wherein the \acrfull{rf}-based techniques have continued to attract most of the research interest for providing high accuracy indoor localization \cite{BiaRapWei:J13}. However, its practical performance for range measurements is greatly degraded in harsh environments due to multipath effects \cite{SeoTan:J08,KulHinGro:C17}, and \acrfull{nlos} conditions \cite{MarGifWymWin:J10}. A further improvement can be achieved with the employment of \acrfull{sri} \cite{MazConAllWin:J18,ConMazBar:J19}. Conventional range-based approaches typically obtain \glspl{de} from measurements, which tend to generate insufficient information for localization. \gls{sri}-based approaches relies on the statistical characterization of the relationship between the inter-node measurements and ranges, which exploit more position-related information and in turns provide more accurate localization.

\Gls{ml} methods are recently introduced to the RF-based localization systems for their ability to accumulate knowledge from data \cite{HuaMolHe:J20}. Such superiority is essential to ill-posed problems where closed-form solutions are complicated or hardly possible to analytically derive, which is often the case in harsh indoor environments. Many recent localization systems have employed \gls{ml} to improve \glspl{de} and subsequently the localization accuracy, such as \cite{VlaEriJavPet:J16}. These methods mostly relies on hand-crafted features.
\Gls{dl} methods take one step further to directly process raw data with high dimensionality. These methods, including
\cite{MaoLinYuShe:C18,LiMazShe:C22}, exploit inherent information and are potential to generate more accurate estimations. However, the aforementioned approaches obtain \glspl{de} rather than \gls{sri}, which offers less information in measurements for localization-related tasks. As a result, integrating the efficiency of \gls{dl} techniques to \gls{sri} generation is promising to fully exploiting the range information in \gls{rf} signals and in turns providing high accuracy localization.

We propose a deep learning approach to generate accurate \gls{sri} directly from received \gls{rf} signals. In the training phase, we train two deep neural modules on a fully labeled database with received signals, propagation conditions, and actual distances. In the testing phase, a unified network composed of such modules can directly generate \gls{sri} from raw signal instances with generalization over different propagation conditions. In addition, The proposed method can conduct \gls{nlos} detection and ranging error mitigation in a single network. Experiments are conducted on two public datasets with \acrfull{uwb} data \cite{WanHuLiLinWanShe:J21}. The results show that the proposed approach can efficiently generate \gls{sri}, and outperforms conventional techniques in terms of both \gls{nlos} detection and range error mitigation.



The remaining sections are organized as follows. Section \ref{sec:model} describes \gls{sri} and the proposed method. Section \ref{sec:network} introduces the according network structure and learning scheme for the proposed method. Performances of the method in different tasks are evaluated with a case study on \gls{uwb} data in Section \ref{sec:exp}. Finally, Section \ref{sec:con} concludes the paper.

\section{Model Formulation}
\label{sec:model}

In this section, we first describe \gls{sri} and then present a three-step learning procedure for \gls{sri} estimation.

\subsection{\gls{sri}-Based Localization System}
\label{sec:problem}

Let $f(\boldsymbol{r}|d)$ be the distribution of waveform measurements $\mathbf{r}$ conditioned on the distance $d$ between a pair of nodes. Following the definition in \cite{MazConAllWin:J18}, the \gls{sri} of a measurements set $\boldsymbol{r}$, denoted $\mathcal{L}_{\boldsymbol{r}}(d)$, is thus any function of distance $d$ proportional to $f(\boldsymbol{r}|d)$, i.e., $\mathcal{L}_{\boldsymbol{r}}(d)\propto f(\boldsymbol{r}|d)$. In addition, $\mathcal{L}_{\boldsymbol{r}}(d)\propto f(d|\boldsymbol{r})$ in absence of prior information on the distance, or using a constant reference prior.

Consider a range-based localization system with each measurements set a collection of a waveform measurement $\mathbf{r}$ and a distance measurement $\bar{d}$. The distance measurement $\bar{d}$ is an instantiation of \begin{equation}  \label{eq:sri}
    \bar{d} = d + \boldsymbol{\mathrm{n}}
\end{equation}
\noindent where $d$ is the distance between a pair of nodes and $\boldsymbol{\mathrm{n}}$ is the measurement noise with PDF given by
\begin{equation}  \label{eq:bias}
    f_{\boldsymbol{\mathrm{n}}}(n)=
\begin{cases}
\mathcal{N}(n;0, \sigma^2_{\text{LOS}})& \text{for LOS cases}  \\
\mathcal{N}(n;b, \sigma^2_{\text{NLOS}})& \text{for NLOS cases}
\end{cases}
\end{equation}
\noindent where $b$ is the positive bias due to propagation condition, i.e., $b_{\text{NLOS}}\geq 0$.

In the following, the propagation conditions are denoted by $\delta$, with $\delta=0$ and $1$ corresponding to \gls{los} and \gls{nlos} conditions, respectively. And the conditional distribution of $\delta$ given $\mathbf{r}$ is $\mathbb{P}(\delta=0|\mathbf{r})$ and $\mathbb{P}(\delta=1|\mathbf{r})$.

The SRI corresponding to a measurement $\mathbf{r}$ described above is
    \begin{equation}
    \label{eq:model}
    \begin{aligned}
        \mathcal{L}_{\mathbf{r}}(d)\propto & ~\mathbb{P}(\delta=0|\mathbf{r})\mathcal{L}_{\text{LOS},\mathbf{r}}(d) + \mathbb{P}(\delta=1|\mathbf{r})\mathcal{L}_{\text{NLOS},\mathbf{r}}(d)  \\
    \end{aligned}
\end{equation}
\noindent with $\mathcal{L}_{\text{LOS},\mathbf{r}}(d)=\mathcal{N}(d;\bar{d},\sigma^2_{\text{LOS}})$ and $\mathcal{L}_{\text{NLOS},\mathbf{r}}(d)=\mathcal{N}(d;\bar{d}-b,\sigma^2_{\text{NLOS}})$.

Such \gls{sri} can generate according DE from measurements by means of the minimum mean square error (MMSE) estimator, by modeling the distance as a random variable (RV). In particular, the DE corresponding to measurement $\mathbf{r}$ is
\begin{equation}
    \label{eq:de}
    \hat{d}=\mathbb{E}\{d|\mathbf{r},\bar{d}\}= \bar{d} - \mathbb{P}(\delta=1|r)b
\end{equation}



\subsection{Learning Procedure}

Given a measurement set of $\mathbf{r}$, the motivation of the presented method is based on two observations: 1) The characteristics of received signals are very different in different propagation conditions, i.e., $f(\bar{d}|\mathbf{r},\delta=0)$ and $f(\bar{d}|\mathbf{r},\delta=1)$ are significantly different in characteristics; 2) The estimation of \gls{sri} $\mathcal{L}_{\mathbf{r}}(d)$ is much harder than the individual estimation of $\mathcal{L}_{\text{LOS},\mathbf{r}}(d)$ and $\mathcal{L}_{\text{NLOS},\mathbf{r}}(d)$. Therefore, we break the estimation of accurate \gls{sri} from measurement $\mathbf{r}$ in three sequential steps:

\begin{enumerate}
    \item The identification step: estimate the propagation condition $\mathbb{P}(\delta=0|\mathbf{r})$ and $\mathbb{P}(\delta=1|\mathbf{r})$;
    \item The estimation step: estimate the \gls{sri} under different propagation conditions, i.e., $\mathcal{L}_{\text{LOS},\mathbf{r}}(d)$ and $\mathcal{L}_{\text{NLOS},\mathbf{r}}(d)$;
    \item Generate \gls{sri} from the estimated distributions via equation\eqref{eq:model}.
\end{enumerate}

To conduct the estimations in these steps, expressions for $\mathbb{P}(\delta|\mathbf{r})$, $\mathcal{L}_{\text{LOS},\mathbf{r}}(d)$ and $\mathcal{L}_{\text{NLOS},\mathbf{r}}(d)$ are required. We adopt deep learning techniques using a fully labeled dataset to approximate these distributions from data.

This paper focuses on propagation conditions, where $\delta\in\{0, 1\}$ refers to the \gls{los} or \gls{nlos} conditions. However, the indicator here can safely scale to more complicated scenarios where $\delta$ denotes more various environmental conditions with a larger set for values, e.g. rooms of different geometries, or different materials of blocking obstacles. Similarly as described in \cite{MazConAllWin:J18,ConMazBar:J19}, the methodology introduced w.r.t range-based measurements can also be used for general measurements related to other positional features such as angle, velocity and acceleration.

\begin{figure}[ht]
      \centerline{
      \includegraphics[width=0.5\textwidth]{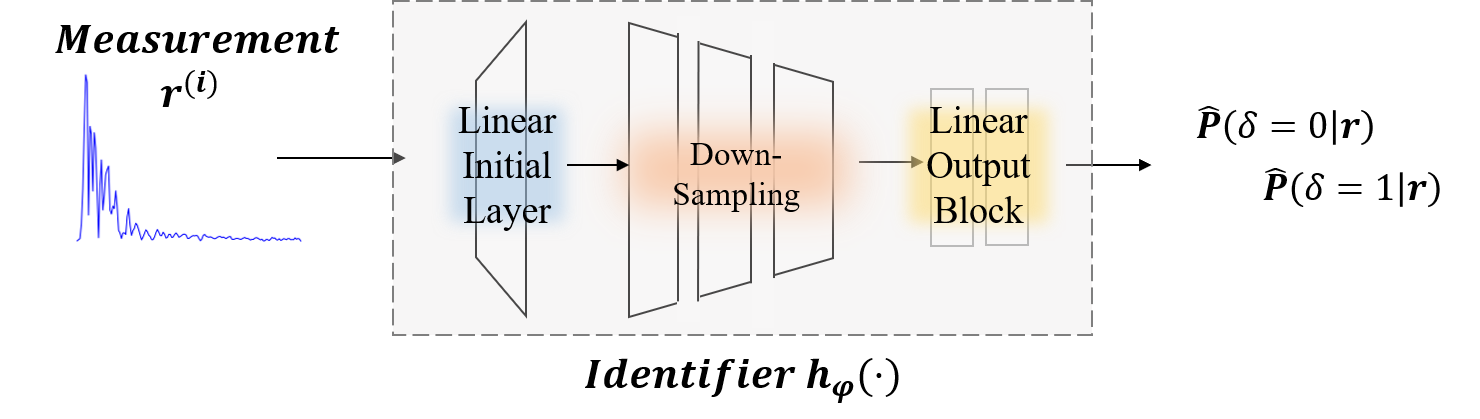}
      }
      \caption{The neural module \textit{Identifier} to infer propagation indicator parameterized by $\boldsymbol{\varphi}$. The module takes measurement as input, and outputs the estimated distribution of propagation indicator, guided by the GT labels from dataset.}
      \label{fig:train_idy}
  \end{figure}
  
  \begin{figure}[ht]
      \centerline{
      \includegraphics[width=0.5\textwidth]{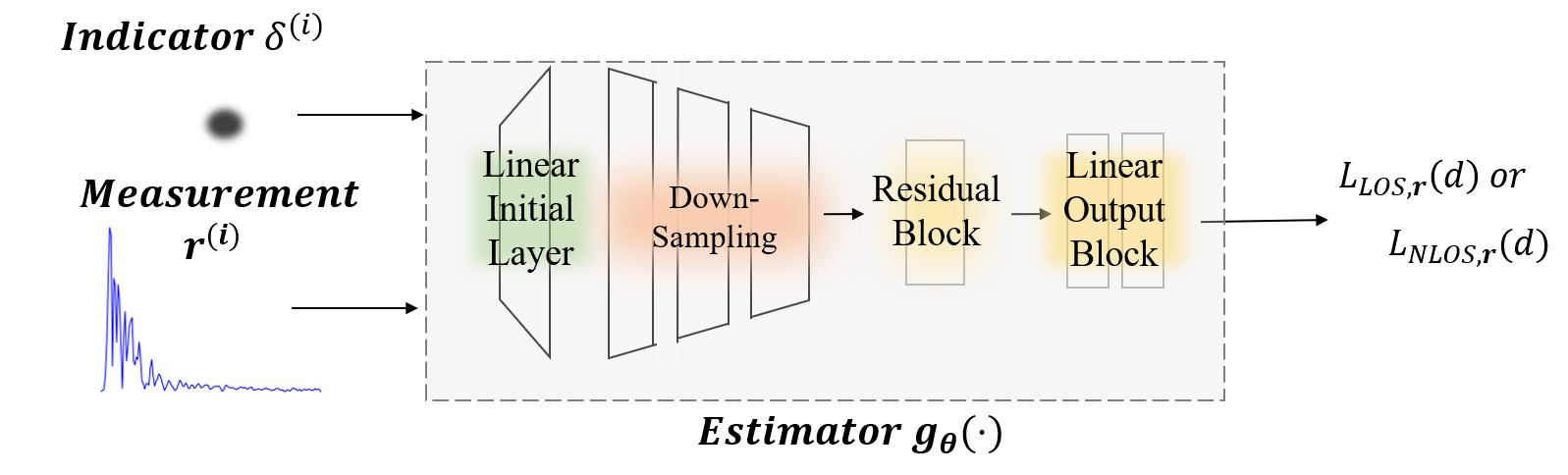}
      }
      \caption{The neural module \textit{Estimator} to infer distance parameterized by $\boldsymbol{\theta}$. The module takes measurements and indicator as inputs, and outputs the estimated distribution of distance, guided by the GT distance from dataset.}
      \label{fig:train_est}
  \end{figure}

\section{Network Implementation}
\label{sec:network}

In this section, we construct neural modules to learn the estimations of target distributions in the aforementioned procedure. 

\subsection{Identifier Module}

Suppose we are given a fully labeled dataset $\mathcal{D}=\{\mathbf{r}^{(i)}, \delta^{(i)}, d^{(i)}\}_{i=1}^N$ with $N$ i.i.d. sample pairs, where $\delta^{(i)}\in\{0, 1\}$ denotes the GT propagation condition, and $d^{(i)}$ denotes the GT distance, both w.r.t. the $i$th measurement $\mathbf{r}^{(i)}$.

We first construct a neural module, referred to as \textit{Identifier}, to conduct the identification step and learn the distribution $\mathbb{P}(\delta|\mathbf{r}^{(i)})$, as illustrated in Fig.~\ref{fig:train_idy}. Specifically, the module learns the mapping $h_{\boldsymbol{\varphi}}(\cdot)$ from $\mathbf{r}$ to $\mathbb{P}(\delta|r)$, taking $\mathbf{r}^{(i)}$ as input and outputting a two-dimensional vector for the estimation of indicator distribution, denoted as $\hat{\mathbb{P}}(\delta=0|\mathbf{r}^{(i)})$ and $\hat{\mathbb{P}}(\delta=1|\mathbf{r}^{(i)})$. The learning of such mapping is guided by a cross-entropy loss of the estimated distribution for the dataset, expressed as:
\begin{equation}
    \label{eq:idy_para}
    \begin{aligned}
    \mathcal{L}_{\text{I}}(\boldsymbol{\varphi};\mathcal{D}) =& \mathcal{L}_{\text{I}}\big(\boldsymbol{\varphi};\{\mathbf{r}^{(i)}, \delta^{(i)}\}_{i=1}^N\big)  \\ 
    =& -\sum_{i=1}^N \boldsymbol{1}_{\delta^{(i)}=0}\log \hat{\mathbb{P}}(\delta=0|\mathbf{r}^{(i)})  \\
    &\phantom{\sum_{i=1}^N-}+ \boldsymbol{1}_{\delta^{(i)}=1}\log \hat{\mathbb{P}}(\delta=1|\mathbf{r}^{(i)})
    \end{aligned}
\end{equation}

\subsection{Estimator Module}

We then construct the neural module for estimation step, referred to as \textit{Estimator}, to learn the \gls{sri} for each indicator, as illustrated in Fig.~\ref{fig:train_est}. Specifically, the module learns the mapping $g_{\boldsymbol{\theta}}(\cdot)$ from $\delta,r$ to $\mathcal{L}_{r}(d)$, generating  $\mathcal{L}_{\text{LOS}}(\mathbf{r}^{(i)})$ if $\delta^{(i)}=0$ and $\mathcal{L}_{\text{NLOS}}(\mathbf{r}^{(i)})$ if $\delta^{(i)}=1$. According to the Gaussian assumption on \gls{sri}, the output of $g_{\boldsymbol{\theta}}(\delta^{(i)}, \mathbf{r}^{(i)})$ is the estimated parameters $\mu^{(i)}, {\sigma^2}^{(i)}$ for the 
associated \gls{sri}. The learning of such mapping is guided by a loss term of the estimated distribution for the dataset, expressed as:
\begin{equation}
    \label{eq:est_para}
    \begin{aligned}
    \mathcal{L}_{\text{E}}(\boldsymbol{\theta};\mathcal{D})=&\mathcal{L}_{\text{E}}\big(\boldsymbol{\theta};\{\mathbf{r}^{(i)}, \delta^{(i)},d^{(i)}\}_{i=1}^N \big)  \\
    =& \operatorname{D}_{\text{KL}}\big(\mathcal{N}(d; \mu^{(i)}, {\sigma^2}^{(i)})\big|\big|\mathcal{N}(d; d^{(i)}, {\epsilon^2}^{(i)})\big) \\
    =& \sum_{i=1}^N \frac{{\sigma^2}^{(i)} + (\mu^{(i)}-d^{(i)})^2}{2\epsilon_0^2} +\log\frac{\epsilon_0}{\sigma^{(i)}} - \frac{1}{2}
    \end{aligned}
\end{equation}
\noindent where $d^{(i)}$ is the GT distance associated with measurement $\mathbf{r}^{(i)}$, $\epsilon_0$ is a small value arbitrarily given by measurement noise in practice.

\subsection{Algorithms}
\label{sec:alg}

The network learning conduct DG-based optimization w.r.t. parameters $\boldsymbol{\varphi}$ and $\boldsymbol{\theta}$. 
During the training phase, network parameters $\boldsymbol{\varphi}$ and $\boldsymbol{\theta}$ are learned separately on dataset $\mathcal{D}=\{\mathbf{r}^{(i)},\delta^{(i)},d^{(i)}\}_{i=1}^N$, with the guidance of loss functions in equations \eqref{eq:idy_para}-\eqref{eq:est_para}.

During the testing phase, the two neural networks work together to generate \gls{sri} from the given measurement. In particular, suppose an instance of measurement $\mathbf{r}$ is given and targeted to generate \gls{sri} from. Such instance is first fed into \textit{Identifier} to get the estimation of $\mathbb{P}(\delta|r)$, e.g., a vector $[\hat{\mathbb{P}}(\delta=0|r), \hat{\mathbb{P}}(\delta=1|r)]$. Then the instance $\mathbf{r}$ together with different propagation indicators $\delta=0$ and $\delta=1$ are fed into \textit{Estimator} to obtain the estimations of $\mathcal{L}_{\text{LOS}}(d)$ and $\mathcal{L}_{\text{LOS}}(d)$, respectively. In particular, instance $\mathbf{r}$ together with $\delta=0$ fed into \textit{Estimator} and generate parameters $\mu_{0},\sigma^2_{0}$, while  with $\delta=1$ generate parameters $\mu_{1},\sigma^2_{1}$. 
According to equation\eqref{eq:model}, \gls{sri} for instance $\mathbf{r}$ is given by
\begin{equation}
\label{eq:sri_t}
    \begin{aligned}
    \mathcal{L}_{\boldsymbol{r}}(d)
    \propto& \hat{\mathbb{P}}(\delta=0|r)\mathcal{N}(d;\mu_{0},\sigma^2_{0}) + \hat{\mathbb{P}}(\delta=1|r)\mathcal{N}(d;\mu_{1},\sigma^2_{1})
    \end{aligned}
\end{equation}

Algorithms \ref{alg:train}-\ref{alg:test} describe the training and testing phases for the proposed algorithm.

\subsection{Evaluation Metrics}
\label{sec:eva}

We adopt evaluation metrics from two aspects for performance evaluation: the accuracy of \gls{nlos} detection, and the accuracy of ranging error estimation. 

After the learning iterations converge, the \gls{nlos} condition can be estimated by maximum-likelihood estimation (MLE) as follows:
\begin{equation}  \label{eq:est_1}
        \hat{\delta} = \arg \max_{\delta}  \hat{\mathbb{P}}(\delta|r)
\end{equation}

The distance can be estimated by MMSE as follows:
\begin{equation}  \label{eq:est_2}
    \begin{aligned}
    \hat{d} = \hat{\mathbb{P}}(\delta=0|r)\mu_{0} + \hat{\mathbb{P}}(\delta=1|r)\mu_{1}
    \end{aligned}
\end{equation}

Given the GT distance $d$ and the measured distance $\bar{d}$ by devices, the estimated ranging error can be achieved by $\hat{b} = \bar{d} - \hat{d}$, and the residual ranging error (remaining error after mitigation) can be $\Delta d = \Vert (\bar{d} - \hat{b}) - d \Vert = \Vert \hat{d} - d\Vert$.


 \begin{algorithm}[!ht]
  \caption{Training Phase}
  \label{alg:train}
  \begin{algorithmic}[1]
    \Require $\mathcal{D}$, the training set, $\alpha$, the learning rate. $m$, the batch size.
    \Require
      $\boldsymbol{\varphi}_0$, initial \textit{Identifier}'s parameters. $\boldsymbol{\theta}_0$, initial \textit{Estimator}'s parameters.
    \While {$\boldsymbol{\varphi}$ has not converged}
        \State Sample $\{\mathbf{r}^{(i)},\delta^{(i)}\}_{i=1}^m \sim \mathcal{D}$ a batch from the dataset.
        \State $h_{\boldsymbol{\varphi}}\leftarrow \nabla_{\boldsymbol{\varphi}}\mathcal{L}_{\text{I}}(\boldsymbol{\varphi};\{\mathbf{r}^{(i)},\delta^{(i)}\}_{i=1}^m)$.
        \State $\boldsymbol{\varphi}\leftarrow\boldsymbol{\varphi} + \alpha * Adam(\boldsymbol{\varphi}, f_{\boldsymbol{\varphi}})$.
    \EndWhile
    \While {$\boldsymbol{\theta}$ has not converged}
        \State Sample $\{\mathbf{r}^{(i)},\delta^{(i)},d^{(i)}\}_{i=1}^m \sim \mathcal{D}$ a batch from the dataset.
        \State $g_{\boldsymbol{\theta}}\leftarrow \nabla_{\boldsymbol{\theta}}\mathcal{L}_{\text{E}}(\boldsymbol{\varphi};\{\mathbf{r}^{(i)},\delta^{(i)},d^{(i)}\}_{i=1}^m)$.
        \State $\boldsymbol{\theta}\leftarrow\boldsymbol{\theta} + \alpha * Adam(\boldsymbol{\theta}, f_{\boldsymbol{\theta}})$.
    \EndWhile
    
    \noindent\Return $\boldsymbol{\varphi}^*$, \textit{Identifier}'s parameter.  $\boldsymbol{\theta}^*$, \textit{Estimator}'s parameter.
  \end{algorithmic}
\end{algorithm}  

 \begin{algorithm}[!ht]
  \caption{Testing Phase}
  \label{alg:test}
  \begin{algorithmic}[1]
    \Require $\mathbf{r}$, the observed signal instance.
    \Require
      $\boldsymbol{\varphi}^*$, \textit{Identifier}'s parameter.  $\boldsymbol{\theta}^*$, \textit{Estimator}'s parameter.
    \State Feed $\mathbf{r}$ to \textit{Identifier} parameterized with $\boldsymbol{\varphi}^*$, obtain $[p_{\text{T}}, 1-p_{\text{T}}]$ and generate distribution $f(\delta|r)$.
    \State Feed $\mathbf{r}$ and label $\delta_{\text{T}}=0$ to \textit{Estimator} with $\boldsymbol{\theta}^*$, obtain $\mu_{\text{T}_0},\sigma^2_{\text{T}_0}$ and generate distribution $f(d|\delta_{\text{T}}=0,r)$.
    \State Feed $\mathbf{r}$ and label $k_{\text{T}}=1$ to \textit{Estimator} with $\boldsymbol{\theta}^*$, obtain $\mu_{\text{T}_1},\sigma^2_{\text{T}_1}$ and generate distribution $f(d|\delta_{\text{T}}=1,r)$.
    \State Generate \gls{sri} of $\mathbf{r}$ via equation\eqref{eq:sri_t}.
    \noindent\Return SRI.
  \end{algorithmic}
\end{algorithm}


\section{Dataset and Implementations}
\label{sec:dataset}

This section utilizes two public UWB datasets utilized for evaluation, and describes the implementation details of our algorithm. The code for our proposed method will be opened to public in the final version.

The methodology presented for \gls{sri} generation is technology-agnostic since it is applicable to any technology capable of providing range-related measurements. This section presents a case study in which ultra-wideband (UWB) signals are employed.

\subsection{Datasets}

We compare the performance of our models with other methods on two public datasets. Both datasets include instances of received UWB measurements, fully labeled with LOS or \gls{nlos} conditions and the actual ranging errors.

\subsubsection{Dataset 1} We use a public dataset from \cite{KleMih:J18} created using SNPN-UWB board with DecaWave DWM1000 UWB pulse radio module. The dataset was generated in two different measurement campaigns in office environments. The first one was recorded in two adjacent office rooms with connecting hallway, including $4800$ measurements in the first room and $5100$ measurements in the second. The second campaign was in a different office environment with multiple rooms, including $25100$ measurements in total. The waveform is represented as the absolute value of CIR, with the length of $152$. 

\subsubsection{Dataset 2} We use a more general public dataset from \cite{zenodo}, created by a campaign with the EVB1000 devices. The dataset consists of $49233$ data samples in total. Each sample includes a CIR waveform of $157$ length, an actual range error, and two environmental labels for the room setting and blocking materials. In particular, measurements are taken in five different room scenarios, including outdoor, big room, medium sized room, small room, and a cross-wall environment. Obstacles of ten different materials that blocking the \gls{los} path are also taken into account.

For both datasets, We utilize $80\%$ of the data samples for training and the rest $20\%$ for testing, as a commonly used strategy for data assignment in deep learning methods.

\subsection{Architecture}
\label{sec:data_arc}

Our framework consists of two sub neural modules for \textit{Identifier} and \textit{Estimator}, as described in Section \ref{sec:model}.
The \textit{Identifier} keeps a simple structure of a initial linear layer, $3$ down-sampling blocks, and a linear output layer. The initial layer concatenate the range feature and received waveform as inputs, and form fused features for the following structures. Each down-sampling block is composed of a down-sampling layer, a ReLU layer, and a dropout layer. They extract environment code with high semantics with low dimensionality, which serves for the predicted condition as well as the side information for \textit{Estimator}.
The \textit{Estimator}, on he other hand, inherits a more delicate structure, with an additional residual block before the output layer.

\subsection{Hyper-Parameters}

We use the Adam \cite{Kingma2015AdamAM} optimizer with $200$ epochs for both datasets. The learning rate is set as $0.0002$, with the decays of first and second momentum of gradients set as $\beta_1=0.9$ and $\beta_2=0.999$ by default, respectively. Other implementation details will be released by the code after final version.
 
The overall model is built in Pytorch, and trained on a GTX $1080$ GPU with a memory of $12$ GB and the accelerator powered by the NVIDA Pascal architecture.

\section{Experiments}
\label{sec:exp}

\begin{figure}[ht]
    \begin{center}
        \subfigure[]{
        \begin{minipage}[t]{0.95\linewidth}
        \centerline{\includegraphics[width=0.9\textwidth]{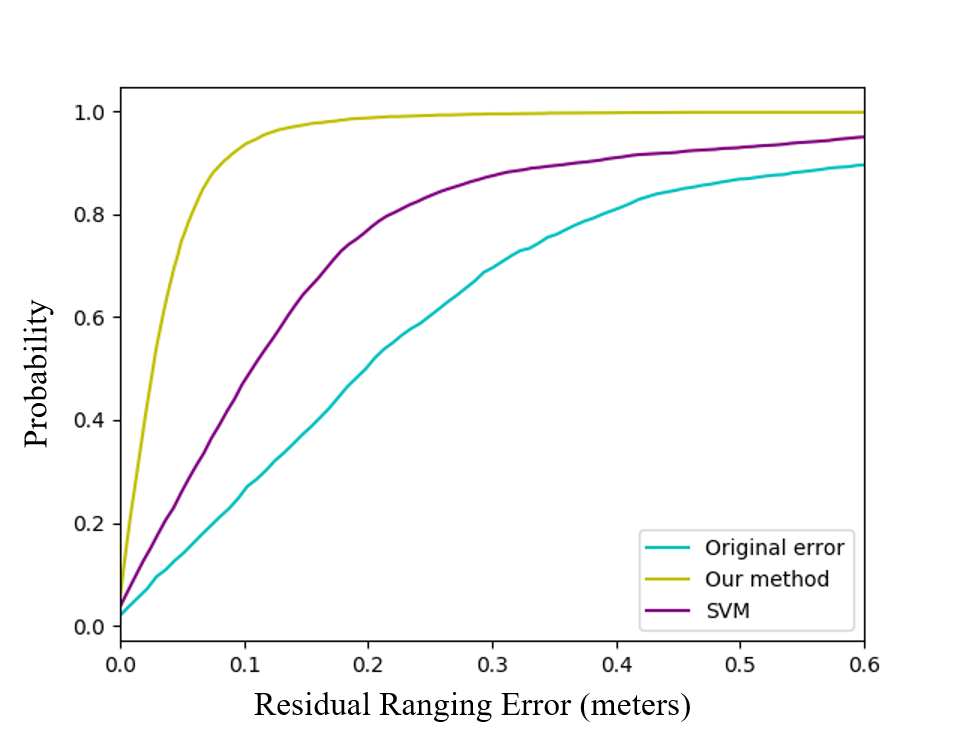}}
        \end{minipage}%
        }  \\
        \vskip -0.08in
        \subfigure[]{
        \begin{minipage}[t]{0.95\linewidth}
        \centerline{\includegraphics[width=0.9\textwidth]{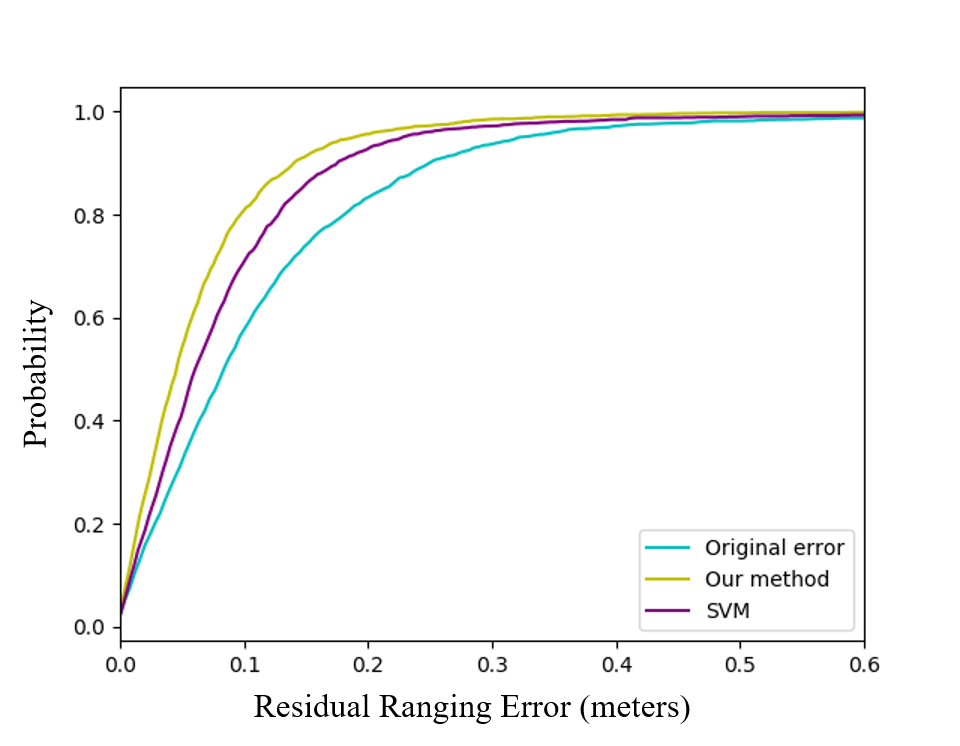}}
        \end{minipage}
        }
        \caption{The CDFs of the residual errors (remaining errors in range measurements after mitigation) after different mitigation methods on (a) \textit{dataset 1}, and (b) \textit{dataset 2}. It can be seen that the proposed method outperforms SVM by a large margin in ranging error mitigation.}
        \label{fig:CDFs}
    \end{center}
\end{figure}

\begin{table*}[t]
\caption{Quantitative comparison in terms of ranging error mitigation on two datasets: Average MAE, RMSE, and Inference Time.}
\label{tab:mitigation}
\begin{center}
\begin{small}
\begin{sc}
\begin{tabular}{l|cc|ccc|ccc}
\toprule
\multirow{2}{*}{\shortstack{Dataset \\Scenarios}} & \multicolumn{2}{c}{Unmitigated} & \multicolumn{3}{c}{SVM-R} & \multicolumn{3}{c}{SRIN} \\
& MAE & RMSE & MAE & RMSE & Time & MAE & RMSE & Time   \\
\midrule
\textit{Dataset 1}     &0.29 &0.44 &0.17 &0.29 &1.91 &\textbf{0.04} &\textbf{0.06} &\textbf{0.06}  \\
\midrule
\textit{Dataset 2}    &0.12 &0.17 &0.09 &0.13 &0.46 &\textbf{0.02} &\textbf{0.05} &\textbf{0.20}  \\
\bottomrule
\end{tabular}
\end{sc}
\end{small}
\end{center}
\end{table*}

\begin{table*}[t]
\caption{Quantitative comparison in terms of NLOS detection on two datasets: Accuracy and Inference Time.}
\label{tab:detection}
\begin{center}
\begin{small}
\begin{sc}
\begin{tabular}{l|cc|cc|cc}
\toprule
\multirow{2}{*}{\shortstack{Environment \\Scenarios}} & \multicolumn{2}{c}{SVM-C} &  \multicolumn{2}{c}{MLP-C} & \multicolumn{2}{c}{SRIN}  \\
& Accuracy & Time & Accuracy & Time & Accuracy & Time  \\
\midrule
\textit{Dataset 1}  &0.820  &1.850 &0.699 &0.550 &\textbf{0.999} &\textbf{0.062}  \\
\midrule
\textit{Dataset 2}  &0.665 &0.654 &0.557 &0.321 &\textbf{0.966} &\textbf{0.201}  \\
\bottomrule
\end{tabular}
\end{sc}
\end{small}
\end{center}
\end{table*}

In this section, we evaluate the proposed method in terms of the performances of \gls{nlos} detection and ranging error mitigation, as described in Sec.\ref{sec:eva}. Both the root mean square error (RMSE) and mean absolute error (MAE) are utilized for the accuracy of ranging error estimation. Quantitative experiments are conducted on the aforementioned datasets, and compared to benchmark methods.

\subsection{Baselines}

Since these methods could not conduct \gls{nlos} detection and error mitigation in a single model, we train separated models for the two tasks. In particular, two different SVMs are trained for either \gls{nlos} classification and ranging error mitigation, referred to as SVM-C and SVM-R respectively, similarly for MLP-C and MLP-R. The proposed network for \gls{sri} is referred to as SRIN for both the \gls{nlos} classification task and ranging error mitigation, since it can conduct both tasks within a unified model. 
The comparison of these learning methods on range error mitigation is in Table \ref{tab:mitigation}, and the comparison on \gls{nlos} detection is illustrated in Table \ref{tab:detection}. Note that both RMSE and MAE are in meters (\textit{m}), inference time per sample is in milliseconds (\textit{ms}), and accuracy is in percentage ($\%$).

\subsection{Performance of Ranging Error Mitigation}
\label{sec:quan_result}

We evaluate the range error mitigation performance in terms of RMSE, MAE, and inference time per sample. Quantitative results on both datasets are presented in Table \ref{tab:mitigation}. The CDFs of the methods on both datasets are shown in Fig.\ref{fig:CDFs}. It can be seen that both the proposed method and SVM conduct effective error mitigation. Note that MLP-R, with performance removed from the table, results in large estimated values and fails to conduct effective mitigation task. By comparison, the proposed approach achieves better results in both datasets, implying both effectivess and generality. Specifically, the proposed method can realize a centimeter-level accuracy, with improvements to SVR of over above $60\%$ for RMSE and $75\%$ for MAE. In addition, the proposed method has the fastest inference speed per sample, indicating the efficiency in practical use.

\subsection{Performance of \gls{nlos} detection}

We evaluate the \gls{nlos} detection performance in terms of classification accuracy of \gls{los} and \gls{nlos} conditions, shown in Table \ref{tab:detection}. It can be seen that all the compared methods can conduct effective \gls{nlos} detection with a good performance over above $55\%$.
The proposed approach shows the best results in both datasets, outperforming SVM and MLP by a large margin. The proposed method also has the faster speed of inference. It is worth noting that the proposed method conduct both error mitigation and \gls{nlos} detection tasks in a unified model, where the compared methods use separately trained models. This further prove the efficiency of the proposed method in practical use.

\section{Conclusion}
\label{sec:con}

We proposed a \gls{dl} approach for \gls{sri} generation in \gls{rf}-based localization systems, and evaluated its performance in terms of \gls{nlos} detection and ranging error mitigation tasks. The proposed approach is implemented by two neural networks, aiming to estimate the distributions of \gls{nlos} condition and range error respectively. The estimated distributions are then combined by a Bayes rule to generate \gls{sri}. 
Experiments on different datasets prove that the proposed method outperforms conventional \gls{ml} methods by a large margin.

\section*{Acknowledgment}
This research is partially supported by the Basic Research Strengthening Program of China (173 Program) (2020-JCJQ-ZD-015-01), the  Basque Government through the ELKARTEK programme, the Spanish Ministry of Science and Innovation through Ramon y Cajal Grant RYC-2016-19383 and Project PID2019-105058GA-I00, and Tsinghua University - OPPO Joint Institute for Mobile Sensing Technology.

\bibliographystyle{IEEEtran}
\bibliography{IEEEabrv,StringDefinitions,SGroupDefinition,refs}

\begin{thebibliography}{10}
\providecommand{\url}[1]{#1}
\csname url@samestyle\endcsname
\providecommand{\newblock}{\relax}
\providecommand{\bibinfo}[2]{#2}
\providecommand{\BIBentrySTDinterwordspacing}{\spaceskip=0pt\relax}
\providecommand{\BIBentryALTinterwordstretchfactor}{4}
\providecommand{\BIBentryALTinterwordspacing}{\spaceskip=\fontdimen2\font plus
\BIBentryALTinterwordstretchfactor\fontdimen3\font minus
  \fontdimen4\font\relax}
\providecommand{\BIBforeignlanguage}[2]{{%
\expandafter\ifx\csname l@#1\endcsname\relax
\typeout{** WARNING: IEEEtran.bst: No hyphenation pattern has been}%
\typeout{** loaded for the language `#1'. Using the pattern for}%
\typeout{** the default language instead.}%
\else
\language=\csname l@#1\endcsname
\fi
#2}}
\providecommand{\BIBdecl}{\relax}
\BIBdecl

\bibitem{BiaRapWei:J13}
O.~Bialer, D.~Raphaeli, and A.~J. Weiss, ``Maximum-likelihood direct position
  estimation in dense multipath,'' \emph{{IEEE} Trans. Veh. Technol.}, vol.~62,
  no.~5, pp. 2069--2079, Jan. 2013.

\bibitem{SeoTan:J08}
C.~K. Seow and S.~Y. Tan, ``Non-line-of-sight localization in multipath
  environments,'' \emph{{IEEE} Trans. Mobile Comput.}, vol.~7, no.~5, pp.
  647--660, May 2008.

\bibitem{KulHinGro:C17}
J.~Kulmer \emph{et~al.}, ``\textnormal{Using DecaWave UWB transceivers for
  high-accuracy multipath-assisted indoor positioning},'' in \emph{Proc. IEEE
  Int. Conf. Commun. Workshop}, Paris, France, May 2017, pp. 1239--1245.

\bibitem{MarGifWymWin:J10}
S.~Maran{\`o}, W.~M. Gifford, H.~Wymeersch, and M.~Z. Win, ``{NLOS}
  identification and mitigation for localization based on {UWB} experimental
  data,'' \emph{{IEEE} J. Sel. Areas Commun.}, vol.~28, no.~7, pp. 1026--1035,
  Sep. 2010.

\bibitem{MazConAllWin:J18}
S.~Mazuelas, A.~Conti, J.~C. Allen, and M.~Z. Win, ``Soft range information for
  network localization,'' \emph{{IEEE} Trans. Signal Process.}, vol.~66,
  no.~12, pp. 3155--3168, Jun. 2018.

\bibitem{ConMazBar:J19}
A.~Conti, S.~Mazuelas, S.~Bartoletti, W.~Lindsey, and M.~Win, ``Soft
  information for localization-of-things,'' \emph{Proc. {IEEE}}, vol. 107, pp.
  2240--2264, Sep. 2019.

\bibitem{HuaMolHe:J20}
C.~Huang, A.~Molisch, R.~He, R.~Wang, P.~Tang, B.~Ai, and Z.~Zhong, ``Machine
  learning-enabled los/nlos identification for mimo systems in dynamic
  environments,'' \emph{{IEEE} Trans. Wireless Commun.}, vol.~19, pp.
  3643--3657, Jan. 2020.

\bibitem{VlaEriJavPet:J16}
S.~Vladimir, L.~E. G., F.~C. Javier, and S.~Peter, ``Kernel methods for
  accurate uwb-based ranging with reduced complexity,'' \emph{{IEEE} Trans.
  Wireless Commun.}, vol.~15, no.~3, pp. 1783--1793, Oct. 2016.

\bibitem{MaoLinYuShe:C18}
C.~Mao, K.~Lin, T.~Yu, and Y.~Shen, ``\textnormal{A probabilistic learning
  approach to UWB ranging error mitigation},'' in \emph{Proc. IEEE Global
  Telecomm. Conf.}, Abu Dhabi, United Arab Emirates, Dec. 2018, pp. 1--6.

\bibitem{LiMazShe:C22}
Y.~Li, S.~Mazuelas, and Y.~Shen, ``\textnormal{Deep Generative Model for
  Simultaneous Range Error Mitigation and Environment Identification},'' in
  \emph{Proc. IEEE Global Telecomm. Conf.}, 2022, \textnormal{To Appear}.

\bibitem{WanHuLiLinWanShe:J21}
T.~Wang, K.~Hu, Z.~Li, K.~Lin, J.~Wang, and Y.~Shen, ``\textnormal{A
  semi-supervised learning approach for UWB ranging error mitigation},''
  \emph{{IEEE} Wireless Commun. Lett.}, vol.~10, no.~3, pp. 688--691, Mar.
  2021.

\bibitem{KleMih:J18}
B.~Klemen and M.~Mihael, ``Improving indoor localization using convolutional
  neural networks on computationally restricted devices,'' \emph{{IEEE}
  Access}, vol.~6, pp. 17\,429--17\,441, Mar. 2018.

\bibitem{zenodo}
S.~Angarano, F.~Salvetti, V.~Mazzia, G.~Fantin, and M.~Chiaberge, ``Deep {UWB}:
  A dataset for uwb ranging error mitigation in indoor environments.'' [OL],
  \url{https://zenodo.org/record/4290069 .X75qYc3-3Dc}.

\bibitem{Kingma2015AdamAM}
D.~P. Kingma and J.~Ba, ``Adam: A method for stochastic optimization,''
  \emph{CoRR}, vol. abs/1412.6980, 2015.

\end{thebibliography}

\end{document}